\documentclass[conference]{IEEEtran}
\IEEEoverridecommandlockouts

\usepackage{cite}
\usepackage{amsmath,amssymb,amsfonts}
\usepackage{algorithmic}
\usepackage{graphicx}
\usepackage{textcomp}
\usepackage{xcolor}
\usepackage{cite}
\usepackage{amsmath,amssymb,amsfonts}
\usepackage{algorithmic}
\usepackage{graphicx}
\usepackage{arydshln}
\usepackage{subfig}
\usepackage{textcomp}
\usepackage{xcolor}
\usepackage{hyperref}
\usepackage{booktabs}
\usepackage[misc]{ifsym}
\usepackage{orcidlink}
\def\BibTeX{{\rm B\kern-.05em{\sc i\kern-.025em b}\kern-.08em
    T\kern-.1667em\lower.7ex\hbox{E}\kern-.125emX}}
\begin{document}

\title{Reliable Mislabel Detection for Video Capsule Endoscopy Data\\
\thanks{This work has been partly funded by the German Federal
Ministry of Research, Technology and Space (BMFTR) in the project MEDGE (16ME0530).}
}

\author{\IEEEauthorblockN{1\textsuperscript{st} Julia Werner}
\IEEEauthorblockA{\textit{Embedded Systems} \\
\textit{University of Tübingen}\\
Tübingen, Germany} \\

\and
\IEEEauthorblockN{2\textsuperscript{nd} Julius Oexle}
\IEEEauthorblockA{\textit{Embedded Systems} \\
\textit{University of Tübingen }\\
Tübingen, Germany}
\and
\IEEEauthorblockN{3\textsuperscript{rd} Oliver Bause}
\IEEEauthorblockA{\textit{Embedded Systems} \\
\textit{University of Tübingen }\\
Tübingen, Germany\\}
\and
\IEEEauthorblockN{4\textsuperscript{th} Maxime Le Floch}
\IEEEauthorblockA{\textit{Department of Medicine I} \\
\textit{University Hospital Dresden}\\
TU Dresden, Germany\\}
\and
\IEEEauthorblockN{5\textsuperscript{th} Franz Brinkmann}
\IEEEauthorblockA{\textit{Department of Medicine I} \\
\textit{University Hospital Dresden}\\
TU Dresden, Germany}
\and
\IEEEauthorblockN{6\textsuperscript{th} Hannah Tolle}
\IEEEauthorblockA{\textit{Department of Medicine I} \\
\textit{University Hospital Dresden}\\
TU Dresden, Germany\\}
\and
\IEEEauthorblockN{7\textsuperscript{th} Jochen Hampe}
\IEEEauthorblockA{\textit{Department of Medicine I} \\
\textit{University Hospital Dresden}\\
TU Dresden, Germany\\}
\and 
\IEEEauthorblockN{8\textsuperscript{th} Oliver Bringmann}
\IEEEauthorblockA{\textit{Embedded Systems} \\
\textit{University of Tübingen}\\
Tübingen, Germany \\}
}


\maketitle
\begin{abstract}
The classification performance of deep neural networks relies strongly on access to large, accurately annotated datasets. In medical imaging, however, obtaining such datasets is particularly challenging since annotations must be provided by specialized physicians, which severely limits the pool of annotators. Furthermore, class boundaries can often be ambiguous or difficult to define which further complicates machine learning-based classification.
In this paper, we want to address this problem and introduce a framework for mislabel detection in medical datasets. This is validated on the two largest, publicly available datasets for Video Capsule Endoscopy, an important imaging procedure for examining the gastrointestinal tract based on a video stream of low-resolution images. In addition, potentially mislabeled samples identified by our pipeline were reviewed and re-annotated by three experienced gastroenterologists.
Our results show that the proposed framework successfully detects incorrectly labeled data and results in an improved anomaly detection performance after cleaning the datasets compared to current baselines. 
\end{abstract}


\section{Introduction}
Proficient performance of machine learning models highly depends on access to large, representative datasets.
However, when large-scale datasets are annotated by humans, the occurence of mislabeled samples is inevitable and a general assumption that all annotations are accurate can introduce substantial issues.
Since deep neural networks tend to overfit on noisy labels, such label noise can adversely affect the generalization and prediction performance~\cite{reed2014training,Tanaka_2018_CVPR}.
In medical applications, the annotation of real-world clinical data is particularly challenging, as it is time-consuming and typically requires specialized clinical expertise, which restricts the pool of qualified annotators.

An example for such medical application is the Video Capsule Endoscopy (VCE), a key diagnostic medical procedure to examine the gastrointestinal (GI) tract, that was first introduced in the early 2000s~\cite{iddan2000wireless,swain2001wireless,thomson2001small}.
The VCE is specifically applied to inspect the small intestine for pathologies while the stomach and colon can be investigated by standard procedures such as a gastroscopy or colonoscopy~\cite{costamagna2002prospective,smedsrud2021kvasir}.
This procedures involves a small pill-sized capsule consisting of a camera, a transmitter, a battery and LEDs, that can be swallowed by patients to record the inside of the digestive tract while it moves through the gastrointestinal organs (mouth, esophagus, stomach, small intestine, colon)~\cite{pillcam,monteiro2016pillcam}.
With current devices on the market, images recorded by the camera are directly transmitted to an on-body receiver for subsequent assessment by gastroenterologists~\cite{pillcam}.
For this application, the long-term objective is on-device anomaly detection in real-time to enable timely diagnosis.

To realize successful screening for pathologies, vision models can be employed, which, however, require access to adequate datasets for effective neural network training.
Importantly, VCE datasets are characterized by a large imbalance with the number of anomalies in the minority~\cite{smedsrud2021kvasir,le2025galar, werner2025enhanced}. In this case, mislabeled data has an even stronger impact and incorrect labels can negatively influence the results strongly. The general presence of low-resolution images obtained in VCEs further complicate evaluations. Additionally, fine-grained annotations can be subjective, as the transition from one class to another can be blurred. To circumvent the potential label ambiguity, one can either increase the robustness of machine learning models or screen for noisy samples and process them prior to neural network training~\cite{patrini2017making,kang2018detection,xiao2015learning}.

\hspace*{2em}\textbf{Our Contribution: } In this work, we developed a framework for mislabel detection in endoscopy datasets and evaluate its performance on the largest, publicly available VCE datasets: Kvasir-Capsule~\cite{smedsrud2021kvasir} and Galar~\cite{le2025galar}. To further assess its reliability, a subset of samples underwent internal review by three co-authors who are experienced gastroenterologists. Two of these authors were also among original authors of the relevant VCE dataset. The review was conducted as part of the collaborative authorship process and served as an expert panel. Incorporating this noise detection pipeline into an anomaly detection workflow leads to substantial improvement compared to the usage of uncleaned datasets and existing baselines. Accordingly, this work proposes an approach for processing medical image datasets before neural network training to boost the overall classification performance.
\section{Related Work}
There are various strategies to handle mislabeled data in the field of machine learning, which focus on increasing the robustness of the models to noisy data during the training process~\cite{patrini2017making}. For instance, instead of adjusting the data points directly, loss functions specifically designed to be less sensitive to erroneous gradients induced by mislabeled samples can be employed~\cite{patrini2017making,zhang2018generalized,wang2019symmetric}.
Additionally, there are methods to clean or filter existing datasets from potential mislabels, e.g. by using ensemble methods~\cite{feng2020label,ostyakov2018label}, or to correct noisy data~\cite{kang2018detection,xiao2015learning}.
Some approaches for reducing label noise are based on the assumption that samples close to each other in a feature space are similar and will therefore have the same label.
Meanwhile, deviating samples with the same label might be considered noisy~\cite{lee2018cleannet,sharma2020noiserank}. 
Confident learning has also been used to determine mislabels by exploiting the confidence of a neural network during training. This technique is based on the assumption that reduced prediction confidence is indicative of mislabeled samples~\cite{northcutt2017learning}. Nevertheless, confidence must be contextualized by the difficulty of classifying each individual sample.

To provide a machine learning-based anomaly detection model for the VCE, large and high quality VCE datasets are essential but remain sparse as they are difficult to be produced and annotated. Recently, the Galar dataset was published, as the largest multi-label VCE datasets~\cite{le2025galar}.
Notably, even with such an extensive and high quality dataset, anomaly detection remains difficult for this application. For example, \cite{le2025galar} achieve a F1-score $5\%$ for polyp detection and $14\%$ for blood detection. While this was improved in later publications to a F1-score of $37.01\%$~\cite{werner2025enhanced} and $54.38\%$~\cite{werner2025seeing} for multiple pathologies pooled together, it emphasizes that there is still room for improvement.
Rather than proposing a new anomaly detection model, this work addresses the underlying issue by identifying noisy data and filtering or correcting mislabeled data points, thereby improving the overall neural network performance.
\section{Methodology}

\begin{figure}[hptb]
    \centering
    \includegraphics[width=\linewidth]{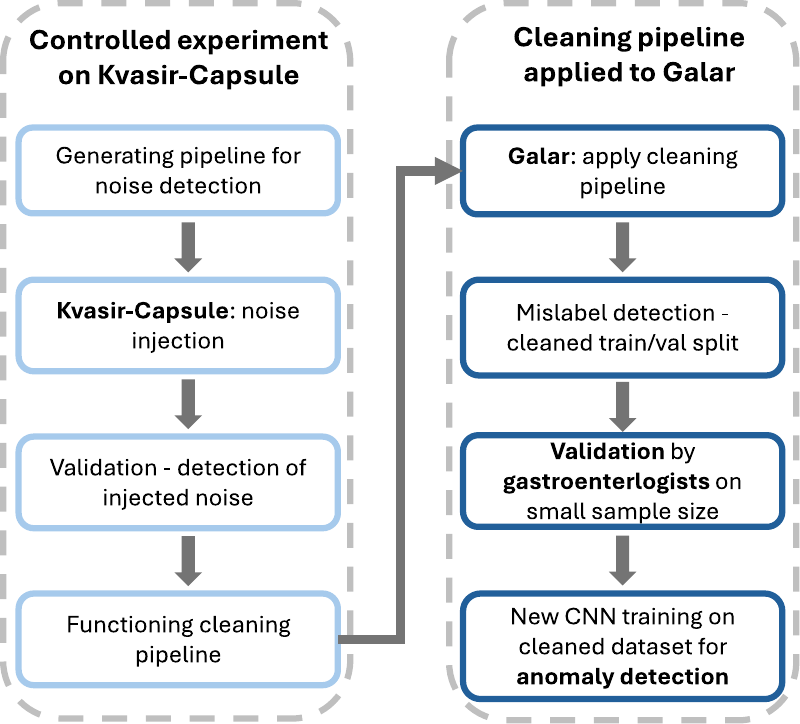}
    \caption{Experimental design: 1) controlled experiment on the Kvasir-Capsule dataset to identify injected noise and 2) mislabel detection on the Galar dataset involving verification by scientific panel of three gastroenterologists with subsequent anomaly detection.}
    \label{fig:experiments_overview}
\end{figure}

One main difficulty when aiming to identify mislabels is a missing ground truth in most datasets.
To address this, the experimental setup of the presented work follows a two-stage design, as illustrated in Figure~\ref{fig:experiments_overview}.
First, a controlled experiment on the Kvasir-Capsule dataset~\cite{smedsrud2021kvasir}  is performed, introducing random label noise (1\%, 5\%, 10\%, 15\%, 20\%) through label flipping. 
Since the falsely labeled data points are known in this experiment, this allows adequate evaluation of the pipeline.

In the second stage, the designed mislabel detection pipeline is applied to the Galar dataset~\cite{le2025galar} with the objective to find pre-existing false labels, resulting in a cleaned version of the dataset.
For validation, $100$ frames identified by our pipeline as potential mislabels were reviewed by a scientific panel of three gastroenterologists and an updated annotation provided.
Finally, a CNN is trained on the cleaned dataset and the anomaly detection performance using the original, unaltered test set evaluated to be comparable to existing baselines.

\subsection{VCE Datasets}
The mislabel detection pipeline was validated on the two largest, publicly available VCE datasets to assess its suitability for wireless capsule endoscopy applications.
The \textit{Kvasir-Capsule} dataset~\cite{smedsrud2021kvasir} contains a total of $47,238$ labeled and $4,694,266$ unlabeled frames.
Consistent with the original publication, the official dataset splits were used to guarantee that images from any given patient were not simultaneously present in the development as well as the test set.

Furthermore, the \textit{Galar} dataset~\cite{le2025galar} was used with the splits generated as in~\cite{werner2025seeing,werner2025enhanced}, which comprises a total of $3,513,539$ labeled images. This multi-label VCE dataset consists of annotated frames with anatomical as well as pathological classes and is intended to be cleaned from noisy data samples in this work.

\subsection{Cleaning Pipeline}
\begin{figure*}[htbp]
  \begin{center}
\includegraphics[width=0.85\linewidth]{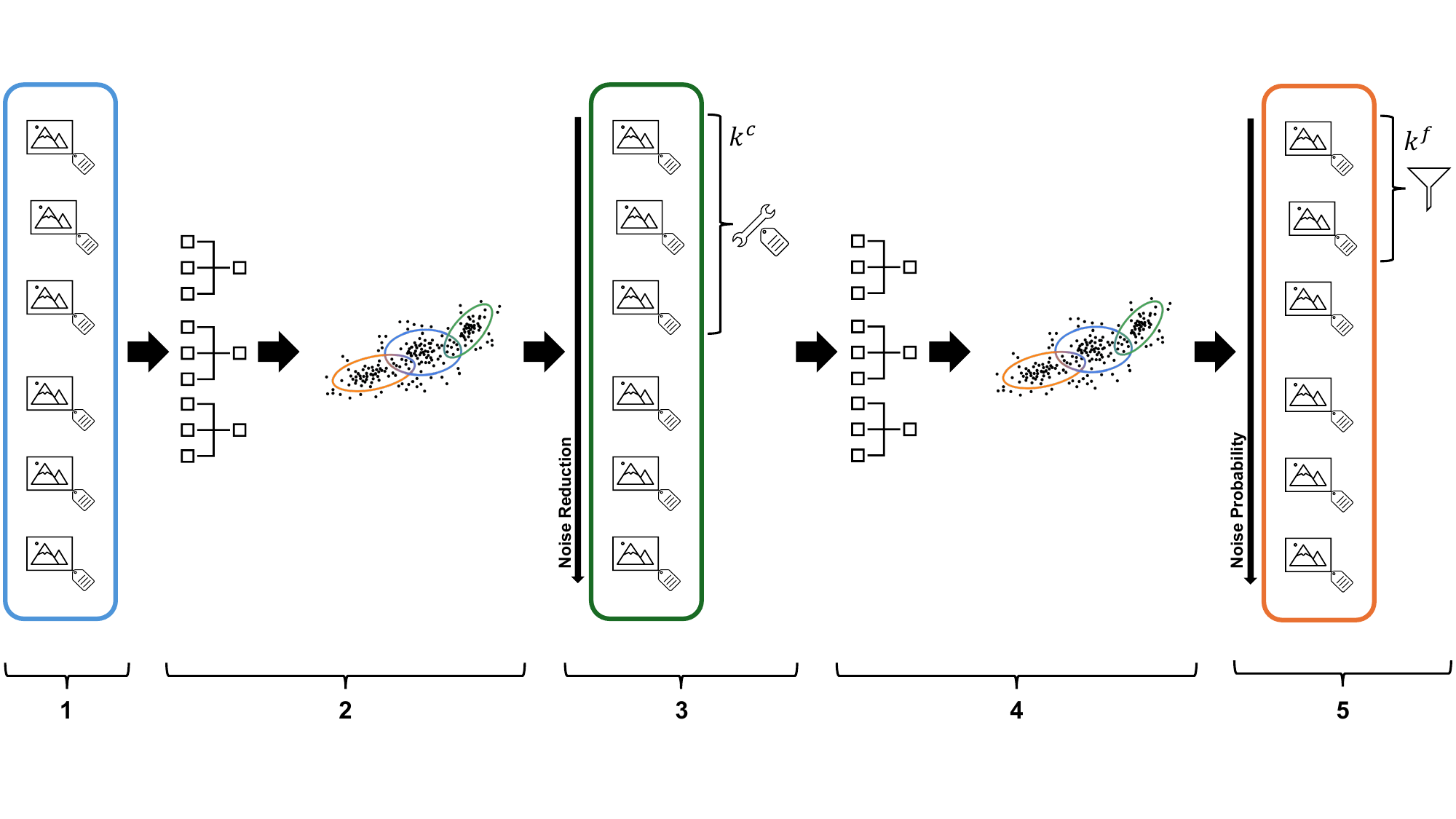}
  \end{center}
  \caption{\label{fig:method} Mislabel correction pipeline: 1. uncleaned dataset, 2. three CNN trainings with subsequent GMM training, 3. correction of $k^c$ labels based on noise reduction, 4. new training to assess the noise probability, 5. filtering $k^f$ mislabeled data samples.}
\end{figure*}

Gaussian Mixture Models (GMMs) have been proven to be a useful tool for the detection of noisy samples~\cite{li2020dividemix}. As they can obtain an increased loss for mislabeled data points compared to correctly labeled samples, they can be used for noise detection in datasets.
Hence, they are an essential component in our experimental setup to detect mislabels as visualized in Figure~\ref{fig:method}.

First, a neural network is trained on the raw dataset in three different processes with subsequent GMM training. 
Based on the prediction and confidence of the neural network, for each sample the probability of having a correct label is determined. This value $p_i^c$ describes the noise probability after a potential correction step for a sample with a noise probability $p_i$. The noise reduction $r_i$ is determined by $p_i - p_i^c$.
Then, the first $k^c$ labels presenting the highest noise reduction are corrected (in a binary case, the labels are flipped). 
Next, three additional CNN and GMM trainings are conducted, concluded with a filtering step in which the first $k^f$ labels with the highest noise probability are filtered to clean the datasets from presumably noisy samples.
To determine whether samples with an increased noise probability should be corrected or filtered out, the concept of \cite{jiang2024more} was applied and the idea of fusing a correction with a filtering step adapted.
In the following, the individual components of the cleaning pipeline are explained in more detail.

\subsection{Noise Injection (Kvasir-Capsule dataset)}
Noise was injected exclusively for the Kvasir-Capsule dataset, for which the true labels are known and serve as ground truth during evaluation. 
For each sample, the average prediction confidence and entropy were computed across all epochs and three independent training runs, normalized, equally weighted, and combined into a single uncertainty score reflecting classification difficulty.
Based on this score, samples were assigned to low-, mid-, and high-uncertainty quantiles. To imitate noisy samples, label noise was then introduced primarily by randomly selecting samples from the mid- and high-uncertainty groups. To preserve the original class distribution, labels were flipped proportionally within each class.

\subsection{CNN Training}
To counteract the influence of the present class imbalances, a focal loss~\cite{lin2017focal} was applied during neural network training. Given that a low model complexity is a key requirement when targeting embedded devices, a MobileNetV3~\cite{howard2019searching} was used, which combine the suitability for low-power embedded devices with a reliable classification performance~\cite{chiu2020mobilenet,howard2017mobilenets}. 
Following \cite{werner2023precise} who used this model type to classify VCE images while considering the total model complexity of 1~M parameters, this network was employed for VCE image classification.

For neural network training, the HANNAH framework~\cite{christoph2022hardware} was employed using a learning rate and weight decay of $1\times 10^{-4}$ for regularization. The models were trained for $15$ epochs if the Galar dataset was used and for $10$ epochs if the significantly smaller Kvasir-Capsule dataset was employed, generally using the AdamW optimizer~\cite{loshchilov2017decoupled}.

\subsection{Gaussian Mixture Model}
A Gaussian Mixture Model (GMM) is a probabilistic model which assumes that all data points of a mixture were generated from a finite number of gaussian distributions~\cite{mclachlan2000finite, permuter2006study}. Each component is represented by a normal distribution, which is defined by a mean $\mu_k$ and a variance $\sigma_k^2$. 
Additionally, each component contains a mixture coefficient $\pi_k$, describing the number of data points belonging to this component. 
The GMM can be trained using the Expectation-Maximization-Algorithm (EM)~\cite{dempster1977maximum}, which consists of an expectation and a maximization step.

Using GMMs, we can identify altered losses for noisy samples compared to correctly labeled samples.
Therefore, within this work, the sklearn implementation of a three-component GMM was used \cite{scikit-learn} and trained with the average loss values of each sample per epoch and process. Subsequently, for each sample it was predicted with which probability it belongs to the gaussian distribution with the highest mean. This probability determines the present noise probability $p_i$ of a sample $i$. The component with the highest mean was determined as the distribution of the wrongly labeled samples and used to compute the noise probability. The component with the lowest mean was determined as the distribution of the samples with correct labels and the third in between those was used as a distribution for difficult samples to learn.

\subsection{Clinical Validation (Galar dataset)}
For the clinical validation of the detected label noise in the Galar dataset, a subset of 100 samples in total was selected for subsequent, internal review by three experienced gastroenterologists who are co-authors 4, 5, and 6 of this manuscript.
Among them, co-authors 4 and 6 were also involved in the original creation of the Galar dataset. 
For the subset selection, all samples were ranked in descending order according to their noise-reduction score and the top 500 images considered for further selection since these images represented the samples with the highest likelihood of being mislabeled.
Finally, a subset of 100 images was chosen, which comprised 70 images depicting normal mucosa and 30 images showing pathological findings to reflect the predominance of normal findings in VCE studies. 
Since a wireless endoscopic capsule may remain at the same position for several minutes without movement, it can sometimes produce visually indistinguishable images.
To ensure a broad coverage as well as variability, images were drawn from more than 50 distinct videos/patients, with a maximum of three samples per video and a minimum separation of 100 frames between samples.
This constraint mitigates redundancy arising from temporally adjacent frames that may capture nearly identical visual content.

\section{Results}

\begin{figure}[hptb]
    \centering
    \includegraphics[width=\linewidth]{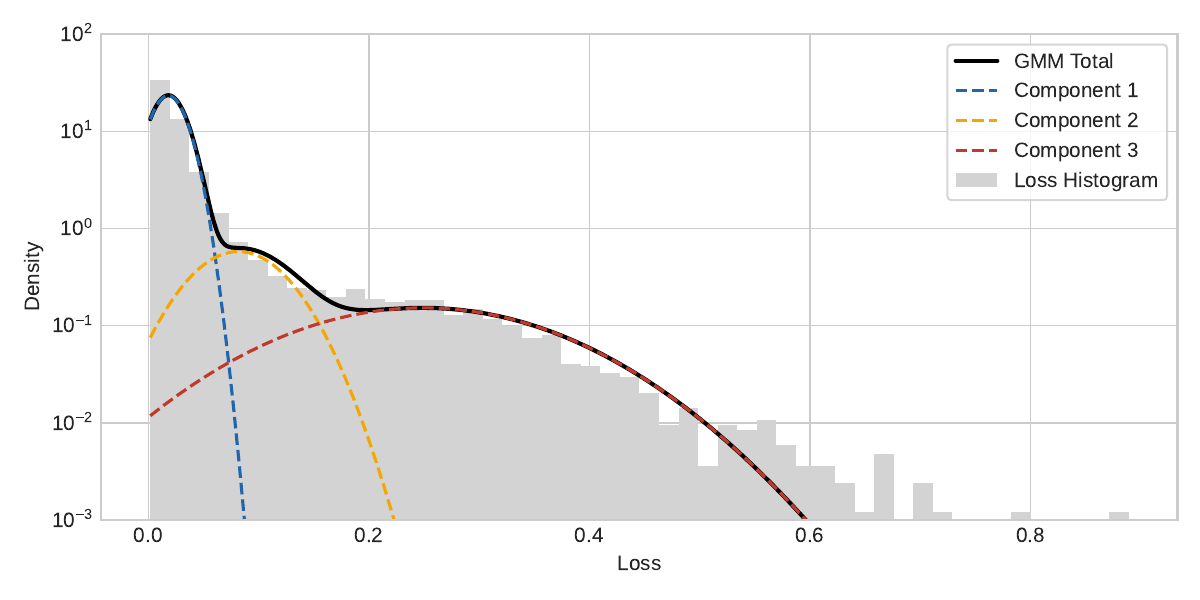}
    \caption{Distribution of the loss values with the GMM total distribution and the individual components (Kvasir-Capsule dataset, 5\% noise injection, correction step).}
    \label{fig:gmm_fit}
\end{figure}

In the following, the results on the Kvasir-Capsule dataset are presented followed by the Galar dataset and finally, the clinical validation by the physicians.
First, using the Kvasir-Capsule dataset, the loss density of the different GMM components is inspected to investigate the differences of the three different clusters within the data samples.
Figure~\ref{fig:gmm_fit} displays the modeling of the GMM during the correction step with 5\% injected noise on the Kvasir-Capsule dataset. This illustrates that the largest number of samples corresponding to correct labels, produces the lowest loss values and is situated below the curve of the first component. The curve of the third component is very flat and contains the highest loss values with strong outliers. For these, it is assumed that the labels are noisy and potentially mislabeled.
This illustrates the prominent differences within the used data and this knowledge can be used for mislabel detection.

For the Kvasir-Capsule dataset, for a defined amount of samples (1\%, 5\% and 10\%), the labels are intentionally flipped randomly and afterwards, the cleaning pipeline used to detect the false/noisy labels. 
Due to this controlled noise injection, after the cleaning pipeline is applied, it can be evaluated proficiently. 

Table~\ref{tab:results_kvasir} presents the mislabel detection results for this experiment and the different amounts of injected noise.
This shows that the majority of mislabeled samples were correctly detected (e.g. 456/471 for 1\% injected noise or 2262/2360 for 5\% injected noise).

\begin{table}[htbp]
	\caption{Cleaning status of samples after injecting 1\%, 5\% or 10\% noise to the Kvasir-Capsule dataset.}
            \begin{center}\resizebox{\columnwidth}{!}{
			\begin{tabular}{lccc}
					\toprule
                    & \multicolumn{3}{c}{Injected Noise }\\
                    Cleaning Status (total number of frames) & 1\% & 5\% &  10\%\\
                    \midrule\midrule
                    Amount of noisy samples & 471 & 2360 & 4722\\
                    Amount of not corrected or filtered samples & 15 & 98 & 367\\
                    Amount of filtered non-noisy samples  & 916 & 975 & 991\\
                    Amount of corrected and/or filtered samples & 456 & 2262 & 4355\\ 
					\bottomrule
				\end{tabular}}
\end{center}
\label{tab:results_kvasir}
\end{table}

\begin{figure}[hptb]
    \centering
    \includegraphics[width=\linewidth]{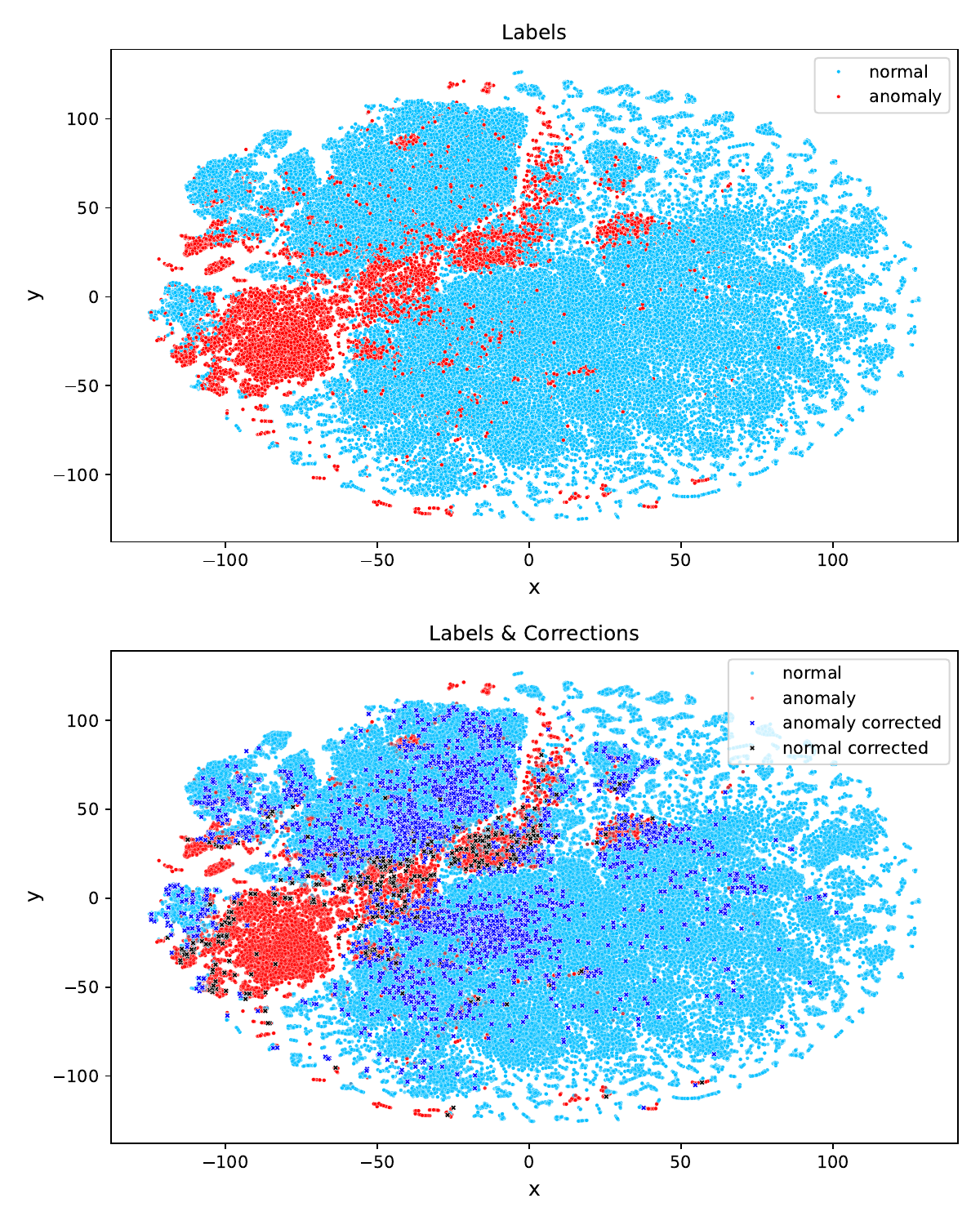}
    \caption{tSNE visualization of the latent representations before (first plot) and after (second plot) mislabel detection with corrected samples indicated (dark blue: anomaly $\xrightarrow{}$ normal, black: normal $\xrightarrow{}$ anomaly).}
    \label{fig:tsne_correction}
\end{figure}

For the Galar dataset, it is first inspected if images of both classes differ in their structure and how well they can be separated. Figure~\ref{fig:tsne_correction} displays t-Distributed Stochastic Neighbor Embedding (t-SNE) plots on the latent space of the samples in the test set in combination with a Principal Component Analysis
(PCA) which reduces the data points to 50 dimensions in total.

\ \\
In the first plot, the samples are colored according to their original annotation in the Galar dataset. A distinct cluster of anomalous samples (red) is visible within the vast majority of normal samples (blue). In addition to this main cluster, individual samples as well as smaller sub-clusters of anomalies occur sporadically within the normal data distribution.
The second plot displays the same latent representations but further visualizes the label corrections conducted by our framework. Dark blue corresponds to samples originally annotated as anomalies, but suspected by our methodologies to represent healthy data. Conversely, black indicates samples which are annotated as normal in the Galar dataset, but our framework classified as most presumably being anomalies.
This demonstrates that our framework primarily corrects labels whose latent representation lie within the cluster of the other class or in transitional regions between both classes.
Consequently, this results in better separated and more coherent clusters.
In total, the cleaning pipeline filtered $167,709$ samples ($4.8\%$ of the Galar dataset) and corrected $31,650$ ($0.9\%$) samples.
The \verb|csv| files containing the dataset splits with the proposed correction or filtering applied as well as a file labeling each sample in the entire Galar dataset as anomalous or healthy, as determined by our annotation pipeline, are available \href{https://figshare.com/s/c39316b5e5d54fbd21f4}{here} for further usage.

\ \\
Since a clearer separation of clusters generally facilitates an improved neural network training and leads to an enhanced classification performance, we subsequently evaluated the classification results on the filtered and cleaned dataset compared to the original, uncleaned dataset and the current baselines.
While conducting the experiments, the cleaned developmental set was used while the test set was kept untouched for better comparability with the existing baselines. However, in the final run, the test set containing the corrected samples was additionally filtered to investigate if this leads to an additional classification enhancement.
The final results are presented in 
Table~\ref{tab:results_galar}.

\begin{table}[htbp]
	\caption{Anomaly detection results on the Galar dataset before and after dataset cleaning compared to existing baselines.}
            \begin{center}\resizebox{\columnwidth}{!}{
			\begin{tabular}{ccccccc}
					\toprule
                    \textbf{Cleaned dev set} & \textbf{Cleaned test set} &  \textbf{Accuracy} & \textbf{F1-Score}  & \textbf{Precision} & \textbf{Sensitivity} &  \textbf{$\varnothing$ max. confidence}\\\midrule
                    Uncleaned\cite{werner2025enhanced} & Uncleaned & 87.28 & 37.01 & 26.59 & 60.88 & -- \\Uncleaned\cite{werner2025seeing} & Uncleaned & 87.7 & 54.38 & 54.14 & 54.69 & --\\
                    \midrule\midrule
                    This work & \\\midrule
                    Uncleaned & Uncleaned & 90.99 & 53.70 & 54.69 & 53.32 & 0.85\\
                    Corrected & Uncleaned & 91.48 & 64.23 & 63.80 & 64.69 & 0.89 \\
                    Filtered & Uncleaned & \textbf{93.83} & \underline{71.58} & \underline{73.03} & \textbf{70.34} & \textbf{0.96}\\
                    Filtered & Filtered & \underline{91.72} & \textbf{73.67} & \textbf{89.88} & \underline{68.05} & \textbf{0.96}\\
					\bottomrule
				\end{tabular}}
\end{center}
\label{tab:results_galar}
\end{table}

It is demonstrated that training a neural network on the corrected/filtered training set leads to substantial classification improvement with a total accuracy of $93.83\%$ and a F1-score of $71.58\%$ compared to uncleaned data with an accuracy of $90.99\%$ and a F1-score of $53.70\%$. Furthermore, a strong advancement can be noted compared to current baselines, which denote F1-scores of $37.01\%$~\cite{werner2025enhanced} and $54.38\%$~\cite{werner2025seeing}.
It is further observed that the model exhibits a boosted confidence if it was trained on the cleaned dataset vs. the uncleaned dataset.
Finally, the precision (89.88\%) is strongly enhanced if the test set is also filtered in addition to the dev set.
This emphasizes the benefit of using the presented cleaning pipeline before conducting anomaly detection.

\ \\
\noindent\textbf{Clinical Validation}\\
Finally, 100 data samples for which our pipeline detected the highest probability of being mislabeled were re-evaluated by a scientific panel of three clinicians. If at least two physicians determined that a label is wrong, this is considered to be the truth in the following. In total, we received a \textbf{Precision@100 score} of \textbf{78}. 
The evaluation is listed in more detailed in the following:
\begin{itemize}
    \item 78\% of suspected labels identified by our pipeline had indeed incorrect labels.
    \item 49\% were originally labeled as normal, but were identified by our pipeline as mislabels and confirmed to be anomalies by the scientific panel.
    \item 29\% were originally labeled as anomaly but were identified by our pipeline as mislabels and confirmed to be indeed normal samples by clinicians.
    \item Only 1\% of anomalies and 21\% of normal data was falsely identified by our pipeline and re-evaluated as correctly labeled by the gastroenterologists.
\end{itemize}

The anomaly detection results in combination with the re-evaluation emphasizes that our cleaning pipeline is well-suited for mislabel identification. In the future, the sample size should be increased to re-evaluate a larger number of suspected samples.

    \begin{figure}[htpb]
		\centering
		\subfloat[I: None - \\  R: Angiectasia]{\includegraphics[width=0.24\linewidth]{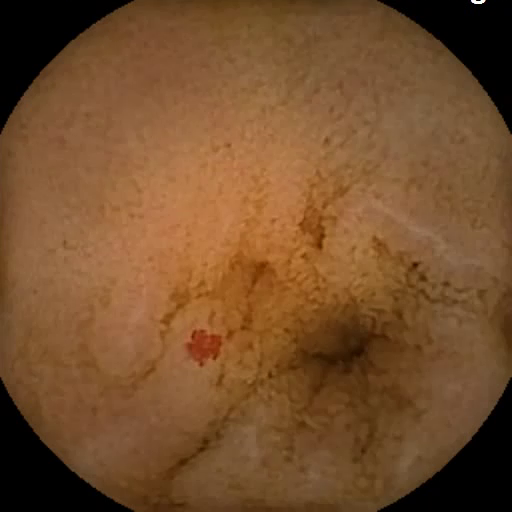}\label{fig:vce_1}}\hspace*{0.1mm}
		\subfloat[I: None - \\ R: Polyp]{\includegraphics[width=0.24\linewidth]{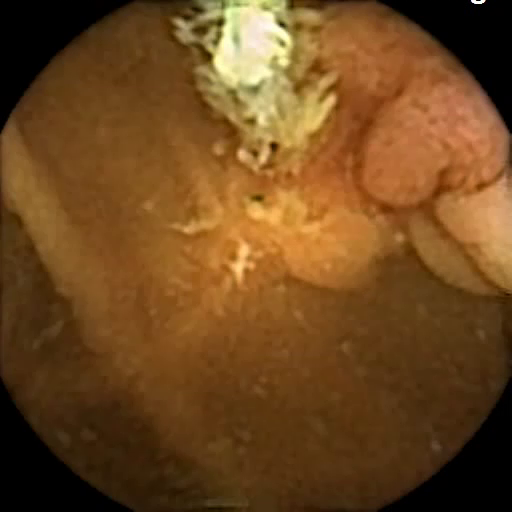}\label{fig:vce_2}}
        \hspace*{0.1mm}
		\subfloat[I: Blood - \\ R: None]{\includegraphics[width=0.24\linewidth]{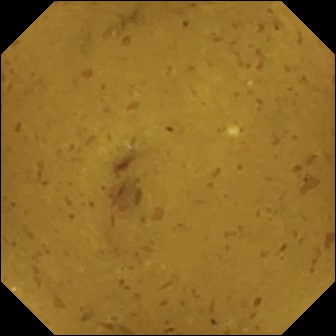}\label{fig:vce_3}}\hspace*{0.1mm}
        \subfloat[I: Polyp - \\ R: None]{\includegraphics[width=0.24\linewidth]{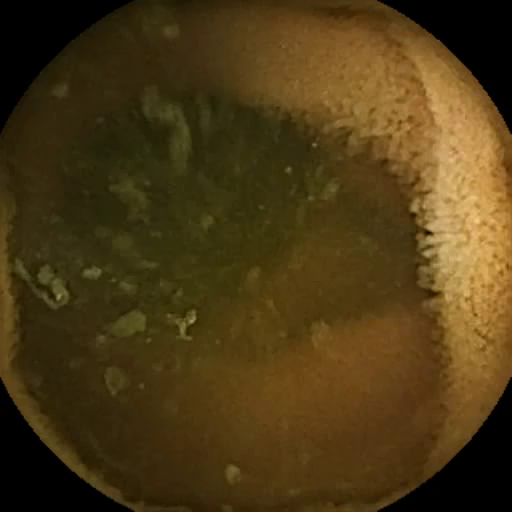}\label{fig:vce_4}}
		\caption{Representative VCE images, that were identified as mislabeled by our pipeline and re-annotated by clinical experts (I: Initial, R: Revised).}
		\label{fig:vce-examples}%
	\end{figure}

Furthermore, Figure~\ref{fig:vce-examples} displays four representative VCE images identified by our proposed approach as mislabeled data, which were subsequently confirmed as incorrectly labeled by the scientific panel and re-annotated accordingly.
The first two images were originally annotated as healthy mucosa, but have been corrected to angiectasia (Figure~\ref{fig:vce_1}) and polyp (Figure~\ref{fig:vce_2}). The second two frames were originally labeled as blood (Figure~\ref{fig:vce_3}) and polyp (Figure~\ref{fig:vce_4}), but were identified as normal samples by our pipeline and then relabeled respectively.
This indicates the effectiveness of our pipeline in detecting present mislabeled data in medical datasets. 
\section{Conclusion}
In this paper, we developed a machine learning-based pipeline to filter and correct medical datasets for potential mislabels. This was validated for the video capsule endoscopy, an essential real-world application to screen for diseases within the gastrointestinal tract.
It was shown that anomaly detection for pathology identification in VCE images achieves superior results if applied to cleaned and filtered datasets, outperforming existing baselines that do not incorporate a prior data cleaning step by obtaining an accuracy of $93.83\%$ and a F1-score of $71.58\%$. 
The effectiveness of this approach was further validated by cross-checking a small sample size by experienced physicians. Given the difficulties reported in previous publications regarding anomaly detection the proposed approach leads to improved results and constitutes a crucial contribution towards enabling on-device anomaly detection during video capsule endoscopies. In future work, all identified noisy samples may be re-annotated by gastroenterologists. Nevertheless, the dataset splits can already directly be used as labeled by our pipeline and are available \href{https://figshare.com/s/c39316b5e5d54fbd21f4}{here}.

\bibliographystyle{splncs04}
\bibliography{bibliography}

\end{document}